%% file: main.tex
\documentclass[pmlr]{jmlr}

\usepackage{longtable}

 \usepackage{booktabs}
 
 \usepackage{color}

\usepackage[load-configurations=version-1]{siunitx} 

\makeatletter
\def\set@curr@file#1{\def\@curr@file{#1}} 
\makeatother

\newcommand{\equal}[1]{{\hypersetup{linkcolor=black}\thanks{#1}}}

\theorembodyfont{\upshape}
\theoremheaderfont{\scshape}
\theorempostheader{:}
\theoremsep{\newline}

\jmlrvolume{[VOLUME \# TBD]}
\jmlryear{2022}
\jmlrworkshop{Machine Learning for Healthcare}

\title{Auto-Gait: Automatic Ataxia Risk Assessment with Computer Vision on Gait Task Videos}

\author{%
\Name{Wasifur Rahman}\equal{Equal contribution} \Email{echowdh2@ur.rochester.edu}\\
\addr Department of Computer Science, University of Rochester
\AND
\Name{Masum Hasan}\footnotemark[1] \Email{m.hasan@rochester.edu}\\
\addr Department of Computer Science, University of Rochester
\AND
\Name{Md Saiful Islam}\equal{Equal contribution} \Email{mislam6@ur.rochester.edu}\\
\addr Department of Computer Science, University of Rochester
\AND
\Name{Titilayo Olubajo}\footnotemark[2] \Email{tolubajo@houstonmethodist.org}\\
\addr Houston Methodist
\AND
\Name{Jeet Thaker}\footnotemark[2] \Email{jthaker@ur.rochester.edu}\\
\addr Department of Computer Science, University of Rochester
\AND
\Name{Abdelrahman Abdelkader} \Email{aabdelka@u.rochester.edu}\\
\addr Department of Computer Science, University of Rochester
\AND
\Name{Phillip Yang} \Email{phil.yang@chet.rochester.edu}\\
\addr Center for Health + Technology, University of Rochester Medical Center
\AND
\Name{Tetsuo Ashizawa} \Email{tashizawa@houstonmethodist.org}\\
\addr Houston Methodist
\AND
\Name{Ehsan Hoque} \Email{mehoque@cs.rochester.edu}\\
\addr Department of Computer Science, University of Rochester
}

\begin{document}

\maketitle
\input{sections/abstract}

\paragraph*{Data and Code Availability}

We will provide an anonymized dataset consisting of extracted pose-features and SARA score upon acquiring approval from the IRB. The code used in this project is available at: \url{https://github.com/Masum06/Automated-Ataxia-Gait}

\input{sections/Introduction}

\input{sections/Related_works}

\input{sections/Dataset_description}
\input{sections/data_processing_feature_extraction}
\input{sections/Models_Results}
\input{sections/Future_work}

\input{sections/conclusion}

\section*{Institutional Review Board (IRB)}
We conducted the experiments with IRB approval of the relevant institution.

\bibliography{sample-base}

\appendix

\input{sections/appendix_selected_features}
%

\end{document}

%% file: sections/abstract.tex
\begin{abstract}


In this paper, we investigated whether we can 1) detect participants with ataxia-specific gait characteristics (risk-prediction), and 2) assess severity of ataxia from gait (severity-assessment). We collected 155 videos from 89 participants, 24 controls and 65 diagnosed with (or are pre-manifest) spinocerebellar ataxias (SCAs), performing the gait task of the Scale for the Assessment and Rating of Ataxia (SARA) from 11 medical sites located in 8 different states in  the United States. We developed a method to separate the participants from their surroundings and constructed several features to capture gait characteristics like step width, step length, swing, stability, speed, etc. Our risk-prediction model achieves 83.06\% accuracy and an 80.23\% F1 score. Similarly, our severity-assessment model achieves a mean absolute error (MAE) score of 0.6225 and a Pearson's correlation coefficient score of 0.7268. Our models still performed competitively when evaluated on data from sites not used during training. Furthermore, through feature importance analysis, we found that our models associate wider steps, decreased walking speed, and  increased instability with greater ataxia severity, which is consistent with previously established clinical knowledge. Our models create possibilities for remote ataxia assessment in non-clinical settings in the future, which could significantly improve accessibility of ataxia care. Furthermore, our underlying dataset was assembled from a geographically diverse cohort, highlighting its potential to further increase equity. The code used in this study is open to the public, and the anonymized body pose landmark dataset could be released upon approval from  our Institutional Review Board (IRB).

\end{abstract}

%% file: sections/Introduction.tex
\section{Introduction}

\label{sec:Intro}
\begin{figure*}
    \centering
    \includegraphics[width=\linewidth]{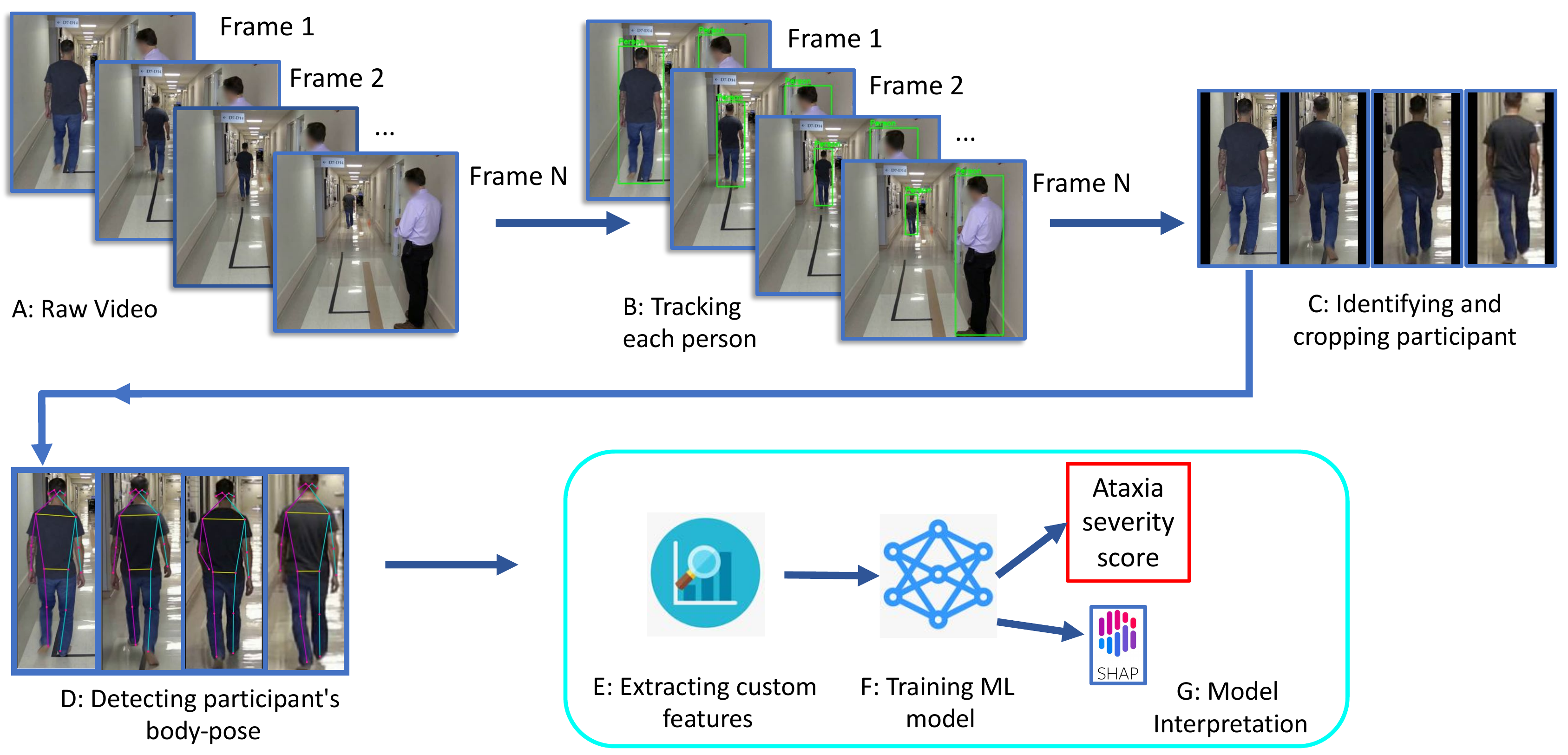}
    \caption{Ataxia risk/severity prediction pipeline from Gait task.
    }
    \label{fig:teaser}
\end{figure*}

Ataxia is a general term referring to a lack of muscle coordination and control, typically arising due to damage to the cerebellum in the brain~\citep{ataxia_stat}. Common symptoms include difficulties walking, deterioration of fine motor skills, imbalance, and trouble eating and swallowing ~\citep{seladi-schulman_2020}. Ataxia may have several different causes including genetics, injuries, stroke, toxins, and more. In this paper, we analyze  two forms of a hereditary ataxia named spinocerebellar ataxia (SCA) types 1 and 3 (SCA1 and SCA3) which demonstrate the fastest progression rate and highest prevalence respectively~\citep{jacobi2012spinocerebellar}.

To date, there are no cures for ataxia; however, early detection and consistent monitoring can greatly improve symptom management and long-term patient prognosis~\citep{ataxia_stat}. The Scale for the Assessment and Rating of Ataxia (SARA) is a popularly used scale for measuring ataxia progression with high inter-rater reliability, inter-rater consistency, and test-retest reliability~\citep{schmitz2006scale,perez2021assessment,hartley2015inter,burk2009comparison}. Although SARA has eight tasks (gait, stance, sitting, speech, etc), we have analyzed the gait task in this paper since 1) SCA heavily impacts common gait characteristics, and 2) gait provides objective, clinically applicable measurements~\citep{buckley2018systematic}.

One particular challenge in applying machine learning to ataxia assessment is constructing a dataset of sufficient size. Given that ataxia is a rare disorder, doing so typically requires collaboration across multiple institutions to accrue enough data. In this study, we have analyzed 155 videos of both ataxia patients and controls collected from 11 different medical sites completing the SARA gait task. However, while the sites all followed the  SARA protocol, each introduced variations to the overall dataset. Figure \ref{fig:gait_walk_different_sites} provides an overview of the variations in data collected from four of these sites. Although this  introduces new challenges in terms of different site settings, lighting conditions, arrangement of doctors and participants, and the presence of other people, it also presents us with a unique opportunity to build a generalizable model. Ataxia is a rare disorder and ataxia specialists are typically clustered in large urban settings. If we can build models that perform reasonably well across different sites, they may be greatly helpful in increasing accessibility of ataxia assessment, especially in underserved communities. 

We designed an automated data processing system (Figure \ref{fig:teaser}) that: a) clips the first six seconds of data of an input video (to normalize height among different settings), b) tracks each person across the video, and c) identifies, crops, and reshapes the participant. Once we obtain cropped images belonging to the participant, we then extract body-pose coordinates, as well as several features designed to measure gait characteristics like swing, stability, speed, etc.\citep{buckley2018systematic}. These features are then fed into a random-forest \citep{breiman2001random} machine learning algorithm to generate predictions for ataxia risk and ataxia severity (descriptions in the next paragraph). Furthermore, we applied SHAP (SHapley Additive exPlanations)~\citep{NIPS2017_7062}, a feature interpretation tool to understand the magnitude and direction of each feature's impact in the model's prediction(s).

Thus, we have addressed two research questions:
\begin{enumerate}
    \item RQ1: Can we provide an early-stage screening of ataxia from the SARA gait task?
    \item RQ2: Can we measure severity of ataxia from the SARA gait task?
\end{enumerate}

In order to answer those questions, we have trained two separate machine learning models. Gait task SARA scores range between 0 and 8, where 0 represents normal gait and 8 represents an inability to walk, even while supported. Figure \ref{fig:num_videos_each_sara_score} provides an overview of how many videos we have corresponding to each gait task SARA score. Since our dataset is focused on early-stage ataxia patients (Sec. \ref{sec:dataset_description}), most of our subjects have gait task SARA scores in the range of [0-3]. In order to answer RQ1, we divided our data into two classes: one with gait task SARA scores of $0$ (Class 0) and another with gait task SARA scores $>0$ (Class 1). Our \textit{Risk-prediction} model can differentiate between these two classes with an 83.06\%  binary accuracy from 10-fold cross-validation (Table \ref{tab:rist-prediction}). To answer RQ2, we divided our dataset into four classes with gait task SARA scores: $0, 1, 2, >2$. For all data with a gait task SARA score $>2$, we replace their label with $3$ since we have very few data with gait task SARA scores $>3$. Our \textit{Severity-assessment} regression model has a mean absolute error of 0.6225 (STD 0.0132) and a  Pearson's correlation coefficient of 0.7268 (STD 0.0144). For both of these models, we ran SHAP analysis. Figure \ref{fig:shap-summary} and Figure \ref{fig:shap-regression-summary}, corresponding to the \textit{Risk-prediction} and \textit{Severity-assessment} models respectively, show that SHAP can provide intuitive, previously-validated results (Sec \ref{sec:model_interpretation}). Furthermore, we ran a   ``leave-one-site-out" analysis, where we exclude data from one site in the training set and test the model's performance on that site's data. This allows us to assess how well the model performs on completely unseen data. While the performance of both models decrease through this method, they still perform significantly better than random chance (Table \ref{tab:rist-prediction}) and  zero-correlation (Table \ref{tab:severity-regression}).

Our contributions are as follows:

\begin{enumerate}
    \item We developed machine learning algorithms for \textit{Risk-prediction} and \textit{Severity-assessment} of ataxia from recorded videos of the SARA gait task. Our models achieve 83.06\% accuracy and an 80.23 F-1 score for the \textit{Risk-prediction} task (Table \ref{tab:rist-prediction}). For the \textit{Severity-assessment} task, we achieve mean absolute error (MAE) of 0.6225 and a Pearson's correlation coefficient of 0.7268 (Table \ref{tab:severity-regression}).
    \item We designed a robust and automated data-processing algorithm that can separate out participants from videos across different site settings that include both doctors and other passersby. Our leave-one-site analysis shows that our models, built around this data-processing algorithm, perform better than random chance across these different sites (Table \ref{tab:rist-prediction} and Table \ref{tab:severity-regression}).
    \item We will release an anonymized version of our dataset containing  body pose coordinates upon acceptance so that the research community can develop better algorithms for ataxia assessment.
\end{enumerate}

%% file: sections/Related_works.tex
\section{Related Works}
Objectively assessing ataxia features is an active area of interest for the researchers. Prior efforts have used sensors to assess ataxia gait and its severity. Other studies have analyzed speech features. There is growing interest in using telemedicine-type platforms to increase accessibility of care for ataxia patients in non-clinical settings. 

\subsection{Detecting Ataxia from Speech}
 Speech anomalies are a common feature of neurological disorders (and may be the earliest indicators of disease) \citep{duffy2019motor}. Thus, recent machine learning-based approaches have analyzed speech recordings, particularly in the domains of Parkinson's disease (PD) and cerebellar ataxia (CA) ~\citep{rahman2021detecting, little2008suitability, kashyap2019automated, kashyap2020automated}. \cite{kashyap2020automated} analyzed recordings of participants repeating single syllables in order to detect and quantify speech-timing anomalies due to CA. 
 \cite{kashyap2019automated} were also successful in distinguishing subjects with CA from healthy controls by tracking vocal tract acoustic biomarkers. To accomplish this, participants recited a tongue-twister phrase (i.e.: ``British Constitution'').

\subsection{Sensors Aiding Automated Analysis of Ataxia}
Wearable sensors and Inertia Measurement Units (IMU) are popular modes for objective diagnosis and assessment of CA. \cite{krishna2019quantitative1} collected features across three tests: finger-to-nose test (FNT), upper limb dysdiadochokinesia test (DDK), and heel-to-shin test (HST) using IMU. IMU can efficiently capture kinematic parameters (for example, acceleration, velocity, angle) and can effectively discriminate between subjects with CA and healthy controls. 
The extracted features using IMU have been found to be highly correlated with well-established clinical assessment scores for measuring severity of CA, including SARA scores \citep{krishna2019limb, krishna2019quantitative}. \cite{nguyen2020entropy} were also successful in using wearable sensors to correctly classify CA and control participants, achieving nearly 80\% accuracy using features extracted from finger and foot tapping task. 
Kinect, a motion sensing device developed by Microsoft, is a well-accepted tool in this research domain. \cite{tran2019automated} combined Kinect with IMU sensors to automatically assess the severity of CA symptoms based on finger chasing and finger tapping tasks. 
\cite{tran2018automated} were also successful in quantifying CA-related disabilities via Kinect as subjects performed the finger chasing task. Kinect is also useful as an alternative to physical therapy, allowing subjects with CA to be rehabilitated from the comfort of their home \citep{ilg2012video, wang2019feature}. 

\subsection{Detecting Ataxia from Gait Task}
Patients with CA may demonstrate clumsy, staggering movements while performing the gait task \citep{serrao2018detecting}. Both wearable sensors and the Kinect system have demonstrated promise in analyzing CA based on subjects performing the SARA gait task \citep{lemoyne2016wearable, phan2019random, honda2020assessment, dostal2021recognition}. \cite{lemoyne2016wearable} gathered accelerometry and gyroscopic data by mounting wireless inertial sensors on the ankle joints of participants. The signals captured were transmitted to a server, where features were then extracted and fed to a machine learning model to distinguish subjects with Friedreich's ataxia from healthy controls. \cite{phan2019random} also employed a similar strategy, but instead of relying on a single gait task, they captured gait features at self-selected slow, preferred and fast walking speeds. The extracted features were successful for both detecting CA and evaluating its severity. \cite{honda2020assessment} analyzed the gait task by extracting features using a Kinect-based system and also produced promising results. Most recently, \cite{dostal2021recognition} obtained accelerometry data using 31 time-synchronized sensors placed on different parts of the body. They obtained an excellent accuracy of 98.0\%, and 98.5\% in classifying between healthy subjects and subjects with CA, using neural networks extracted features from the shoulders and head/spine respectively. Another recent, similar approach obtained 95.8\% accuracy in classifying Ataxic and normal gait by using deep learning to analyze frequency components of accelerometric signals simultaneously recorded at multiple body positions. While the above mentioned studies depend on using external sensors, \cite{ortells2018vision} explored automated video assessment to analyze Ataxic gait. However, the dataset used in the study was simulated by 10 healthy volunteers, so it did not contain real data from subjects with Ataxia. 

\subsection{Application of Teleneurology}

Recently, researchers have tried to increase the use of telemedicine to care for patients with CA, which is also recommended by experts \citep{manto2020medical}. Notably, \cite{grobe2021development} developed $\text{SARA}^{\text{home}}$, a video-based instrument to measure ataxia severity from a subject's home. Independently, a user performs 5 of the 8 SARA items; items were modified so that the tool could be easily self-applied without the presence of a provider. After all the tasks are completed, the recorded videos are securely transferred to be evaluated by an expert rater offline using SARA criteria. 
While this tool represents an advance in remote delivery of care, it still requires a human rater, even if asynchronously. Separately, ~\cite{summa2020development} created SaraHome, a tool that interlinks the Microsoft Kinect 2.0 and Leap Motion Controller (LMC), and launched a pilot study to gauge participant reception and feasibility of the tool.
The tool was well-received; however, 
to date, it has not been tested on actual video footage to assess its performance compared to traditional measures. 

\subsection{Automated Video Analysis}
While automated video analysis is challenging, it has the potential to increase accessibility of care. \cite{jaroensri2017video} introduced an automated method for quantifying motion impairment of ataxia patients by analyzing video recordings of the Finger to Nose Test (FNT). They used a neural network for pose estimation and optical flow techniques to track participant hand motion. Their models performed comparably to traditional clinical assessments of ataxia severity.

In this paper, we use machine learning to distinguish between ataxia gait and normal gait, as well as assess the severity of ataxia gait, from recorded videos of subjects performing the SARA gait task in clinical settings. We introduced 14 features that can be computed without sensors; these features are reflective of well-established ataxia gait characteristics. In addition, we combined data across multiple clinical sites to validate the effectiveness of the proposed features across different environments.

%% file: sections/Dataset_description.tex
\section{Dataset Description}
\label{sec:dataset_description}

\begin{table*}[h!]
 \caption{Demographics information of our participants 
 }
  \label{tab:demographics}
\begin{tabular}{|l|l|l|l|l|}
\hline
 Status & Number of participants (N=89) & Mean age (std deviation) & \# Female (percentage) & \#Videos \\ \hline
PwA     & 65                        & 47.07(9.08)              & 36 (55.38)             & 116     \\ \hline
Control       & 24                        & 41.73(8.72)              & 15 (62.5)              & 39       \\ \hline
\end{tabular}
\end{table*}

\begin{figure*}[h!]
  \centering
  \includegraphics[width=\linewidth]{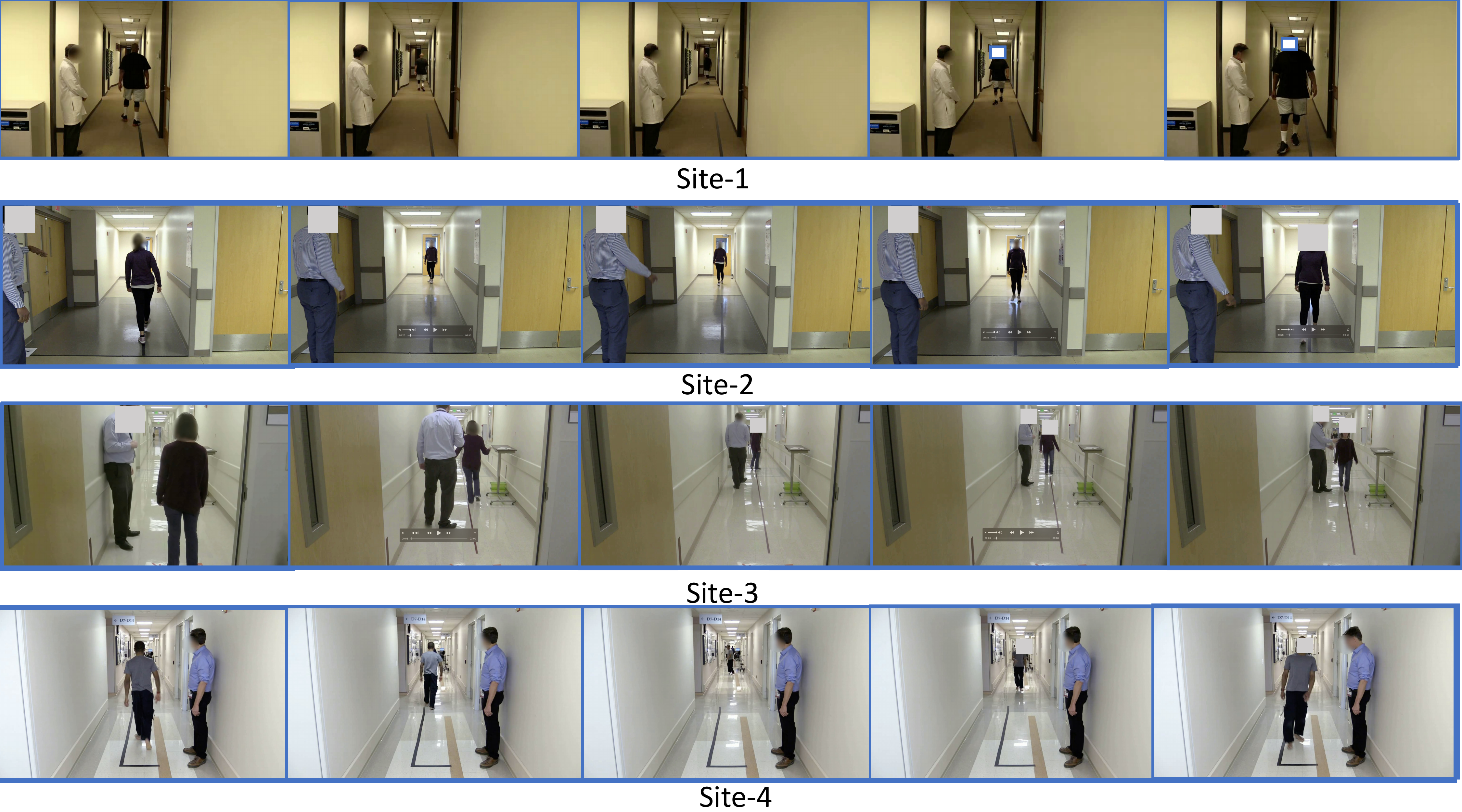}
  \caption{Gait walk sample from four sites. We analyzed the first six seconds of data for each video. In all cases, the participants was still going away from the camera after six seconds (corresponding to first two images for each site). }
  \label{fig:gait_walk_different_sites}
\end{figure*}

\begin{figure}[h!]
  \centering
  \includegraphics[width=\linewidth]{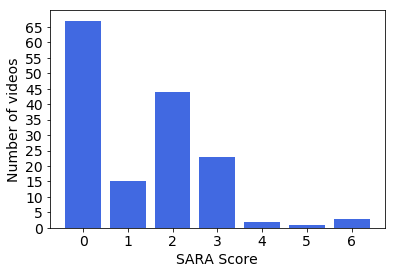}
  \caption{Number of videos for each SARA score}
  \label{fig:num_videos_each_sara_score}
\end{figure}

\begin{figure}[h!]
  \centering
  \includegraphics[width=\linewidth]{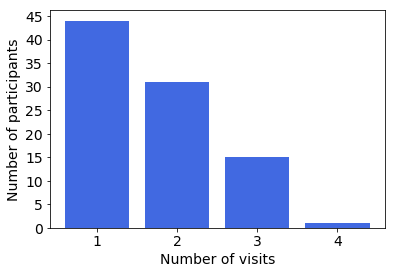}
  \caption{Number of participants  who completed 1, 2, 3, or 4 visits.}
  \label{fig:num_part_num_visit}
\end{figure}

\begin{figure}[h!]
  \centering
  \includegraphics[width=\linewidth]{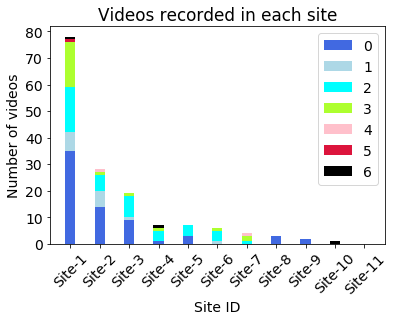}
  \caption{Number of videos recorded at each site and  gait-SARA score [0-6] mix in those videos}
  \label{fig:video_dist_each_site}
\end{figure}

Our dataset consists of recorded videos of 89 unique participants completing SARA tasks at annual visits in a clinical setting across eleven different medical sites in USA. Of these 89 individuals, 24 were healthy controls. The remaining 65 were either already symptomatic with spinocerebellar ataxia type 1 or type 3 (SCA1 and SCA3 respectively) or were otherwise pre-symptomatic. We will refer to these 65 participants as People-with-Ataxia (PwA). Across the 65 PwA and 24 controls, we obtained 116 and 39 unique videos  from each group respectively. Table \ref{tab:demographics} provides demographic information on all 89 participants.
If any one of the following exclusionary criteria were met, then a participant was deemed ineligible to participate in the study: 
\begin{itemize}
    \item Known genotype consistent with that of other inherited ataxia (for example, Freidrich's ataxia) 
    \item Concomitant disorder(s) that affected assessment of ataxia severity during study 
    \item Changes in physical and occupational ataxia therapy in the two months prior to study enrollment 
\end{itemize}

Not all sites provided equal amounts of data. Figure \ref{fig:video_dist_each_site} provides an overview of how many videos were recorded at each site and the gait-specific SARA score distribution of them. As can be seen, the first three sites supplied most of the videos and have a good mix of videos with different SARA scores. Figure \ref{fig:gait_walk_different_sites} provides snapshots of videos collected at the various locations. While SARA was administered at each site, there nonetheless were some site-specific variations present in the videos, including: 

\begin{enumerate}
    \item Each site had different lighting conditions
    \item Providers would sometimes walk beside the participants (Site 3) and thus overlap with the participants
    \item Providers would either stand to the left (Site 2) or to the right (Site 4) of the participants
    \item Because some videos were captured in high-traffic areas, other people (not participants or providers) would occasionally be captured in the recordings (Site 4)
\end{enumerate}

Furthermore, this study is a multi-visit study, meaning that some of the captured videos are of the same participant and collected over an extended time period.  Figure \ref{fig:num_part_num_visit} provides the distribution of number of study visits completed across all participants.


While the videos include footage of the 89 participants completing all eight tasks in SARA, we specifically focused on the gait task portions. As described previously, SARA assigns a score from 0 to 8 for the gait task, where 0 represents normal walking and 8 represents inability to walk, even with support.
Figure ~\ref{fig:num_videos_each_sara_score} provides a breakdown of the distribution of gait-specific SARA scores across all 155 collected videos (PwA and control). As can be seen, the majority of participants had gait-specific SARA scores between 0 and 3. These SARA scores were provided by multiple neurologists with high inter-rater agreement (on average, the score from different raters had a difference of $\pm 0.5$ on same video).

%% file: sections/data_processing_feature_extraction.tex
\section{Data Processing and Feature Extraction}

\begin{figure}[]
  \centering
  \includegraphics[width=\linewidth]{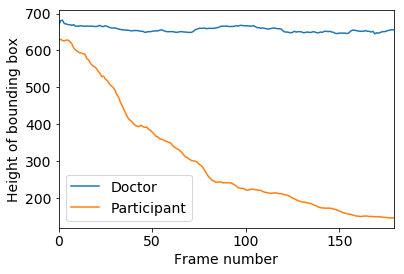}
  \caption{Change of bounding box height for doctor and participant in a sample video (from the first 6 seconds of video)}
  \label{fig:heuristics_height_change}
\end{figure}

\subsection{Data Prepossessing}
\label{sec:patient-huristics}
For each recorded video, we analyzed the first six seconds of data. Since participants took different amounts of time to complete the gait task, we restricted analysis to the first six seconds in order to normalize the data. During those six seconds, all participants were moving away from the camera and did not have time to turn around and return back towards the camera. 
As can be seen in Figure \ref{fig:gait_walk_different_sites}, the videos contained noise, including: 

\begin{enumerate}
    \item Doctors evaluating participants, who are captured in the video. They typically were either at rest or moved a few steps. 
    \item Passersby captured in the video. They typically moved horizontally and were visible only for a few frames.
\end{enumerate}

To isolate the participant's movement, we had to separate the participant from the surrounding environment. To accomplish that, we designed appropriate heuristics after making the following observations:

\begin{enumerate}
    \item At every site, the camera was statically located behind the participant and the participant walked  away from the camera. 
    \item The participant was visible in almost every frame and constantly moving away from the camera. Thus, the participant continuously got smaller in size. 

\end{enumerate}

To isolate the participant's gait, we processed the videos through the following three steps (Figure \ref{fig:teaser}). First, we detected the presence of each individual in the video. Second, we tracked each individual through a bounding-box, and third, designed heuristics to correctly isolate the participant. 

We used a Faster R-CNN object detection model ~\citep{ren2015faster} to process all video frames. This model outputs a box that assigns boundaries to each object in the frame. With at least 80\% probability, we could successfully separate objects detected as "people". Then, we applied the SORT algorithm to track the identified persons throughout the video ~\citep{bewley2016simple}. The SORT algorithm tracks the objects detected via the Faster R-CNN model by assigning a unique ID to each one. Thus, we could generate a series of bounding boxes for all persons detected in the videos. 

As mentioned previously, the participants consistently decreased in size throughout the duration of the videos. Figure \ref{fig:heuristics_height_change} provides an example of how the height of the bounding boxes containing the doctor and the participant changed over time. By representing height in the $i$th frame as $H_i$, we can calculate a score for height change,
\begin{equation}
    score = \sum_{i=0}^{N-1} (H_i - H_{i+1})
\end{equation}

We calculated a $score$ for each of the persons tracked; the individual with the highest score has walked the farthest distance consistently, hence, they are the participant. Afterwards, we manually inspected all the data to ensure the correctness of this method. An additional, unintended benefit of this simple method is that it filtered out many of the extraneous passersby. Because these individuals only appear in a small number of frames, their $scores$ will be low.  

\subsection{Feature Extraction}

\begin{table*}[]

\caption{Description of features.  Since we initially extract a sequence of features for each Feature group entry,  we calculate five sub-features to represent that sequence: mean, std-deviation, max, min, range (max-min), entropy. The values 1 and 2 in Source column represent \citep{buckley2018systematic} and \citep{ honda2020assessment} respectively. }
  \label{tab:feature_list}
\begin{tabular}{|l|l|l|l|}
\hline
Feature group                                 & Measurement                                                                                            & Estimation for                           & Source \\ \hline
feet\_dist                                & Distance between feet                                                                             & Step width, step length, swing           & 1, 2   \\ \hline
feet\_angle                                & Angle between the x-axis and  line joining feet               & Swing, step length, swing                & 1      \\ \hline
right\_x  & X-Coordinate of right feet                                                                     & Step length, step width, swing           & 1, 2   \\ \hline

right\_y  & Y-Coordinate of right feet                                                                     & Step length, step width, swing           & 1, 2   \\ \hline

left\_x & X-Coordinate of left  feet                                                                     & Step length, step width, swing           & 1, 2   \\ \hline

left\_y & Y-Coordinate of left  feet                                                                     & Step length, step width, swing           & 1, 2   \\ \hline

left\_knee\_bent      & Angle between the left thigh and  knee                                                         & Swing                                    & 1      \\ \hline

right\_knee\_bent       & Angle  between the right thigh and  knee                                                         & Swing                                    & 1      \\ \hline

imbalance                                  & Horizontal deviation of the mid-shoulders and hip. & Stability                                & 2      \\ \hline
tilt                                        & Angle of shoulder line with horizontal axis                                                            & Stability                                & 2      \\ \hline
nose\_x                          & X-coordinate of the nose                                                                                & Stability                                & 2      \\ \hline

nose\_y                          & Y-coordinate of the nose                                                                                & Stability                                & 2      \\ \hline

stance                                      & \% of frames where the participant was stationary                                              & Stance phase, swing phase                & 1      \\ \hline
height\_reduction                           & Total height reduction for the 6s walk                                                    &  Speed & 1      \\ \hline
\end{tabular}
\end{table*}

After separating out the frames containing the participant of interest, we extracted several features based on previously validated gait characteristics of SCA patients ~\citep{buckley2018systematic,honda2020assessment}. Notably, these features reflect common ataxia gait features including step width, step length, swing, stability, stance phase, swing phase, and speed.

Table \ref{tab:feature_list} provides a short description of our features. For first 13  entries listed in the ``Feature group" column in Table \ref{tab:feature_list}, we identified six sub-features: mean, std-deviation, max, min, range ($max-min$), entropy. Since we get a sequence of data for these 13 entries, we need to compute some sample statistics to capture their information succinctly. For the last entry height\_reduction was itself a number (Section \ref{sec:patient-huristics}), hence no sub-feature was necessary. Thus, we finally have $13*6 +1 = 79$ features to build our models. From which, we select the final features using Recursive Feature Elimination with Cross-Validation (RFECV) algorithm.

We derived entropy features as they can capture the irregularity in a sequence of data. Mathematically, a periodic signal will have zero entropy, and a random signal will have high entropy. As SCA patients have difficulties in maintaining a consistent rhythm in their walk, entropy-related features should effectively capture these irregularities in walking pattern. We used a open-source Python implementation \citep{spectral_entropy} of the spectral entropy method originally designed to capture irregularities in EEG brain signals \citep{inouye1991quantification}.






%% file: sections/Models_Results.tex
\section{Models and Results}


\begin{table}[]
\caption{Ataxia risk prediction  (binary classification) model performance. The model was evaluated in 20 iteration of 10-fold cross-validation and leave-one-site-out manner (Section \ref{sec:risk-prediction}). 
}
\label{tab:rist-prediction}
\begin{tabular}{|l|rr|rr|}
\hline
\multicolumn{1}{|c|}{Test} & \multicolumn{2}{l|}{F1 score (\%)}	& \multicolumn{2}{l|}{Accuracy (\%)}	\\ \cline{2-5} 
\multicolumn{1}{|c|}{}	& \multicolumn{1}{c|}{Mean}	& \multicolumn{1}{c|}{STD} & \multicolumn{1}{c|}{Mean}	& \multicolumn{1}{c|}{STD} \\ \hline
10-fold CV	& \multicolumn{1}{r|}{80.23} & 9.19	& \multicolumn{1}{r|}{83.06} & 6.79	\\ \hline
Site 2	& \multicolumn{1}{r|}{80.5}	& 2.65	& \multicolumn{1}{r|}{80.54} & 2.64	\\ \hline
Site 3	& \multicolumn{1}{r|}{65.34} & 5.04	& \multicolumn{1}{r|}{65.79} & 4.85	\\ \hline
Site 4	& \multicolumn{1}{r|}{74.65} & 6.32	& \multicolumn{1}{r|}{81.43} & 6.55	\\ \hline
Site 5	& \multicolumn{1}{r|}{66.94} & 5.83	& \multicolumn{1}{r|}{72.86} & 4.29	\\ \hline
Site 6	& \multicolumn{1}{r|}{94.55} & 16.36	& \multicolumn{1}{r|}{98.33} & 5.0	\\ \hline
\end{tabular}
\end{table}

\begin{table}[]
\caption{Ataxia severity prediction  (regression between 0$\sim$3) model performance. The model is evaluated with 20 iteration of 10-fold cross-validation and leave-one-site-out manner. In the second case, a model is tested on a site that was excluded during training (Section \ref{sec:severity-prediction}). The mean and standard deviation of Mean Absolute Error (MAE) (lower better) and Pearson's Correlation Coefficient (higher better) are reported. 
}
\label{tab:severity-regression}
\begin{tabular}{|l|rr|rr|}
\hline
\multicolumn{1}{|c|}{Test} & \multicolumn{2}{c|}{MAE}                                  & \multicolumn{2}{c|}{Corr Coeff.}                          \\ \cline{2-5} 
\multicolumn{1}{|c|}{}                      & \multicolumn{1}{c|}{Mean}   & \multicolumn{1}{c|}{STD}    & \multicolumn{1}{c|}{Mean}   & \multicolumn{1}{c|}{STD}    \\ \hline
10-fold CV                                  & \multicolumn{1}{r|}{0.6225} & \multicolumn{1}{c|}{0.0132} & \multicolumn{1}{r|}{0.7268} & \multicolumn{1}{c|}{0.0144} \\ \hline
Site 2 & \multicolumn{1}{r|}{0.5673} & 0.0 & \multicolumn{1}{r|}{0.7081} & 0.0 \\ \hline
Site 3 & \multicolumn{1}{r|}{0.7723} & 0.0 & \multicolumn{1}{r|}{0.4569} & 0.0 \\ \hline
Site 4 & \multicolumn{1}{r|}{1.0886} & 0.0 & \multicolumn{1}{r|}{0.2146} & 0.0 \\ \hline
Site 5 & \multicolumn{1}{r|}{0.6334} & 0.0 & \multicolumn{1}{r|}{0.6244} & 0.0 \\ \hline
Site 6 & \multicolumn{1}{r|}{0.4402} & 0.0 & \multicolumn{1}{r|}{0.4297} & 0.0 \\ \hline
\end{tabular}
\end{table}





\begin{figure*}[htbp]
\floatconts
  {fig:scatter_regression}
  {\caption{Features' impact on models prediction (computer by SHAP)}}
  {%
    \subfigure[Risk prediction model (binary)]{\label{fig:shap-summary}%
      \includegraphics[width=0.47\linewidth]{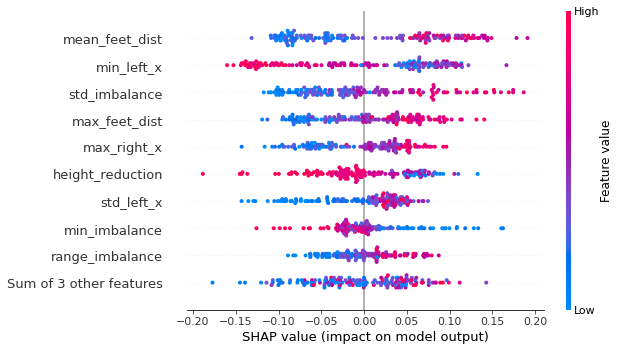}}%
    \qquad
    \subfigure[Severity assessment model (regression)]{\label{fig:shap-regression-summary}%
      \includegraphics[width=0.47\linewidth]{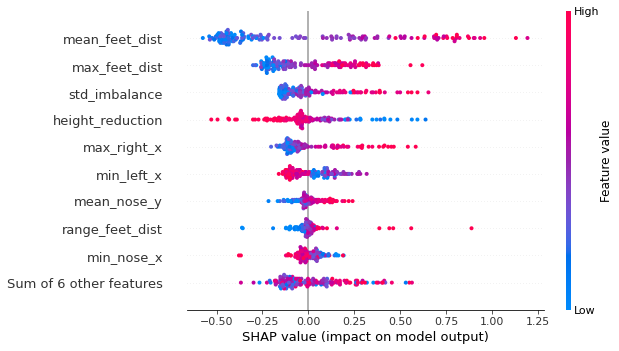}}
  }
\end{figure*}

\begin{figure*}[htbp]
\floatconts
  {fig:scatter_regression}
  {\caption{Visualizing the predictions for Severity-assessment task. The red line in Fig. \ref{fig:scatter} denotes  ideal scatter plot. The light-green color in Fig. \ref{fig:kernel_density} shows an estimation of the prediction density. }}
  {%
    \subfigure[Scatter plot]{\label{fig:scatter}%
      \includegraphics[width=0.47\linewidth]{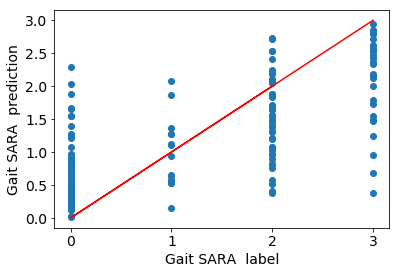}}%
    \qquad
    \subfigure[Gaussian kernel density estimation]{\label{fig:kernel_density}%
      \includegraphics[width=0.47\linewidth]{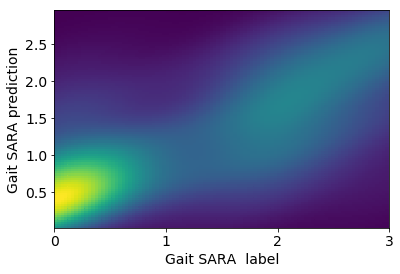}}
  }
\end{figure*}

Once we computed the numeric features from the participants' gait data, we trained two different machine learning classifiers to answer RQ1 (Risk-prediction) and RQ2 (Severity-assessment) posed in Section \ref{sec:Intro} .

\subsection{Risk-prediction}
\label{sec:risk-prediction}

First, we trained a random forest classifier to classify participants at risk for ataxia from healthy controls. All participants with a $>0$ score  are classified as `at risk' (Class 1) and healthy controls are considered 'risk-free' (Class 0). To select the optimal set of features for training this model, we employed the recursive feature elimination with cross-validation method~\citep{recursive_feature_selection}. The complete list of features used for training this model is available in \ref{apd:risk_pred_selected_features}.

There are 88 videos in Class 1 and 67 videos in Class 0. Therefore, the accuracy of choosing the majority class is 0.568. Hence, accuracy of more than 0.568 indicates the model is performing better than random chance. We evaluated the model using 10-fold cross-validation and leave-one-site-out methodologies. In the second case, we trained the model on all sites but one and tested the model on the held-out site. Good performance on this task provides a useful metric for evaluating our model's generalizability in completely unseen situations. We perform all of the experiments 20 times to minimize randomness. The mean and standard deviation of the 20 runs are reported in Table \ref{tab:rist-prediction}. We can see from Table \ref{tab:rist-prediction} that the model achieves an F1 score of 80.23\% and an accuracy of 83.06\% in 10-fold cross-validation. Clinically which means, the model is able to distinguish a CA patient from non-CA subject with above 80\% correctness. The low standard deviation between results indicates that the results are fairly consistent between different runs. In leave-one-site-out tests, the model performs fairly well, despite never seeing data from a particular site.

\subsection{Severity-assessment}
\label{sec:severity-prediction}
Second, in order to predict the SARA score directly from gait data, we designed a random forest regression model. As our primary motivation is to identify early stage CA, we focus on lower scores of SARA and categorize all SARA scores of 3 or above to simply 3. The model takes the numeric features from participants walking as input and predicts a continuous score between 0 and 3. We use the feature importance score of the random forest algorithm to identify the best features for training this model (a complete list is available in \ref{apd:severity_assess_selected_features}.

We evaluate the model via calculating the mean absolute error (MAE) and a Pearson's correlation coefficient (PCC). A lower MAE is better, as it indicates the mean difference between the model's prediction and the actual SARA score. PCCs range from -1 to 1, where higher values are better, implying a correlation between the prediction and ground truth. On 10-fold-cross-validation, our model achieves MAE and PCC scores of 0.6225 and 0.7268 respectively. We can see from Table \ref{tab:severity-regression} that in most cases, the model achieves less than 1.0 MAE through the leave-one-site analysis and its predictions are positively correlated with ground truth SARA scores. The low standard deviation indicates that the predictions are consistent across runs. By manually inspecting 6 datapoints (Figure \ref{fig:gait_walk_different_sites}) in Site 6, we find that although the model has high accuracy in Table \ref{tab:rist-prediction} and the lowest of all MAE in Table \ref{tab:severity-regression}, a single misclassification results in a relatively lower but still acceptable correlation score. Figures \ref{fig:scatter} and \ref{fig:kernel_density} visualize the positive correlation through a scatter plot and Gaussian kernel density estimation.

\subsection{Model Interpretation}
\label{sec:model_interpretation}
Figure \ref{fig:shap-summary} and \ref{fig:shap-regression-summary} give us insights into which features most influence our models' predictions. We can see that the coordinates of the left and right feet, as well as the distance of steps, play important roles in model prediction. Based on these summaries, we see that wider steps (\texttt{min\_left\_x, max\_feet\_dist, max\_right\_x}, etc.) and variation in steps (\texttt{std\_left\_x, range\_feet\_dist}, etc.) both affect ataxia scores; this is consistent with \citep{buckley2018systematic, schmitz2006scale, classification}, who also found that step length, step width, step distance, and differences in step length between the left and right feet are indicative of CA.
\citep{buckley2018systematic, schmitz2006scale} also reported that individuals with CA have lower walking speeds than healthy controls. In our analysis, \texttt{height\_reduction} is a direct reflection of walking speed, as reduced height indicates a further distance traveled, which is proportional to distance traveled in a fixed duration (6 seconds in this study). We can see that in both graphs, \texttt{height\_reduction} negatively affects CA score. In addition, \citep{schmitz2006scale} found that ataxia participants' have oscillatory walking patterns, which is effectively captured through the \texttt{imbalance} features in our model; \texttt{range\_imbalance, std\_imbalance, min\_imbalance}, etc.) contribute to higher SARA scores, thus denoting more advanced ataxia. Position of the nose (Figure \ref{fig:shap-regression-summary}) also can be indicative of oscillation in upper body movement. Overall, Figure \ref{fig:shap-summary} and \ref{fig:shap-regression-summary} demonstrate that the most significant features in our model are consistent with those identified in prior literature. 

%% file: sections/Future_work.tex
\section{Limitations and Future Work}

\subsection{Incorporating Additional Characteristics}
In this study, the classification features are computed with clever engineering grounded on existing literature. However, a more exhaustive study of ataxia and its effects could reveal more discriminative features that can result in better performance.
In order to normalize different walking distance between sites, we only considered the first 6 second of walking forward part of video-recorded gait. However, in a standard setting, a patient walks forward, turns and then walks back to the initial position. Taking the full walk into consideration could potentially provide more information. Further work can also be done in tandem walk, Stance task, finger to nose task, etc. using similar techniques.
In the future, we plan to find novel ways to capture features reflective of different task in Cerebellar Ataxia prediction and cover a larger part of SARA score metric.

\subsection{Limitations of dataset}
Our dataset was collected on people either diagnosed or at-risk of Cerebellar Ataxia (SCA). While SARA can be also used to assess other types of ataxia aside from SCAs -- such as ataxia due to brain tumors and ataxia telangiectasia ~\citep{perez2021assessment} -- the target populations nonetheless differ. Therefore, our results may not translate to other other types of Ataxia. In addition, the videos were collected in clinical settings, and all the participants were assisted by a doctor. It would be interesting to explore whether our findings hold for home-recorded videos, potentially without any supervision of another person.



\subsection{Differentiating between Parkinson's disease and ataxia}
At least 15\% of patients diagnosed with Parkinson's disease (PD) are later found to be incorrectly diagnosed ~\citep{schrag2002valid}. SCAs and other types of ataxia present very similar symptoms to PD, namely slowed movements, tremors, stiffness, and difficulty balancing ~\citep{park2015parkinsonism}. In the words of an ataxia patient, \textit{``I know there are people walking around with exactly what I have and they don’t know it.... It is often misdiagnosed because it resembles other things like multiple sclerosis, stroke and Parkinson’s disease.”}\citep{ataxia_misdiagnosis}. Both SARA for measuring ataxia severity and the Movement Disorder Society Unified Parkinson's Disease Rating Scale (MDS-UPDRS) include a gait task \citep{goetz2008movement,subramony2007sara}. In collaboration with researchers, we have \textit{already amassed} a collection of videos of PD participants completing the gait task of the MDS-UPDRS. We intend to build machine learning models that can distinguish between PD and ataxia in individuals from gait video footage.

\subsection{Creating a platform for automated, end-to-end SARA scoring for clinic and home use}

While our current models only score the gait task section of SARA from captured videos, this work nonetheless opens up possibilities for development of an end-to-end framework that provides scores for all eight tasks based on recorded footage.
Not only could such a system be utilized in the clinic, but in the home as well. 
In the future, we are interested in developing a platform that collects video footage of users completing a modified version of SARA in a home environment \citep{grobe2021development} and scores each section accordingly (perhaps through a smartphone application or web application).  
This would improve accessibility of valuable neurological information and may inform earlier interventions.
In the future, we plan to develop an analogous scoring scheme using machine learning-based approaches to complement SARA results.

%% file: sections/conclusion.tex
\section{Conclusion}
In this paper, we present machine learning models that can detect participants displaying ataxia-impacted gait characteristics and can measure severity of ataxia from the SARA gait task. Our models open up new possibilities for the future development of a user-friendly, automated platform that can remotely assess ataxia in non-clinical settings. This would have various applications, including independent use by patients and use by stakeholders in patient-centric clinical trials. Moreover, the platform can be re-purposed to detect whether elderly individuals are at risk of falls, since a gait speed of less than 1 meter/second strongly predicts falls~\citep{kyrdalen2019associations}. As falls are one of the main reasons for fractures, physical incapacity, and even death \citep{terroso2014physical} in the elderly, our tool can be helpful in limiting fall-related accidents and providing timely care to this vulnerable population. 

%% file: sections/appendix_selected_features.tex
\section{Selected Features}
\label{apd:selected_features}
\subsection{Features Used For Risk-prediction task}
\label{apd:risk_pred_selected_features}
 [mean\_feet\_dist, std\_left\_x, std\_imbalance, std\_nose\_y, max\_feet\_dist, max\_right\_x, min\_left\_x, min\_imbalance, range\_left\_x, range\_imbalance, range\_nose\_x, height
\_reduction]

\subsection{Features Used For Severity-assessment task:}
\label{apd:severity_assess_selected_features}
[
'mean\_feet\_dist', 'height\_reduction', 'max\_feet\_dist', 'max\_right\_x', 'std\_imbalance', 'min\_left\_x', 'range\_imbalance', 'range\_nose\_x', 'mean\_nose\_y', 'min\_nose\_x', 'entropy\_nose\_y', 'std\_tilt', 'std\_left\_x', 'range\_feet\_dist', 'range\_left\_x'
]
